\documentclass[runningheads,a4paper]{llncs}

\usepackage{amssymb}
\usepackage{graphicx}
\usepackage{cite}
\usepackage{url}
\newcommand{\keywords}[1]{\par\addvspace\baselineskip
\noindent\keywordname\enspace\ignorespaces#1}

\begin{document}

\mainmatter  

\title{Virtual PET Images from CT Data Using Deep Convolutional Networks: Initial Results}

%
%
%
\author{Avi Ben-Cohen$^{1}$%
\and Eyal Klang$^{2}$ \and Stephen P. Raskin$^{2}$ \and Michal Marianne Amitai$^{2}$\and\\
Hayit Greenspan$^{1}$}

\authorrunning{Ben-Cohen et al.}
\titlerunning{Virtual PET Images from CT Using Deep Learning}%



\institute{$^{1}$Tel Aviv University, Faculty of Engineering, Department of Biomedical Engineering, Medical Image Processing Laboratory, Tel Aviv 69978, Israel\\
$^{2}$Sheba Medical Center, Diagnostic Imaging Department, Abdominal Imaging Unit, affiliated to Sackler school of medicine Tel Aviv University, Tel Hashomer 52621, Israel}

%
%
\maketitle

\begin{abstract}
In this work we present a novel system for PET estimation using CT scans. We explore the use of fully convolutional networks (FCN) and conditional generative adversarial networks (GAN) to export PET data from CT data. Our dataset includes 25 pairs of PET and CT scans where 17 were used for training and 8 for testing. The system was tested for detection of malignant tumors in the liver region. Initial results look promising showing high detection performance with a TPR of 92.3\% and FPR of 0.25 per case. Future work entails expansion of the current system to the entire body using a much larger dataset. Such a system can be used for tumor detection and drug treatment evaluation in a CT-only environment instead of the expansive and radioactive PET-CT scan.

\end{abstract}

\section{Introduction}
The combination of positron emission tomography (PET) and computerized tomography (CT) scanners have become a standard component of diagnosis and staging in oncology \cite{Weber,Kelloff}. An increased accumulation of Fluoro-D-glucose (FDG), used in PET, relative to normal tissue is a useful marker for many cancers and can help in detection and localization of malignant tumors \cite{Kelloff}. Additionally, PET/CT imaging is becoming an important evaluation tool for new drug therapies \cite{Weber2}. 
Although PET imaging has many advantages, it has a few disadvantages that make it a difficult treatment to receive. The radioactive component can be of risk for pregnant or breast feeding patients. Moreover, PET is a relatively new medical procedure that can be expensive. Hence, it is still not offered in the majority of medical centers in the world. 
The difficulty in providing PET imaging as part of a treatment raises the need for an alternative, less expensive, fast, and easy to use PET-like imaging. In this work we explore a virtual PET module that uses information from the CT data to estimate PET-like images with an emphasis on malignant lesions. To achieve the virtual PET we use advanced deep learning techniques with both fully convolutional networks and conditional adversarial networks as described in the following subsections.

\subsection{Fully Convolutional Networks}
In recent years, deep learning has become a dominant research topic in numerous fields. Specifically, Convolutional Neural Networks (CNN) have been used for many challenges in computer vision. CNN obtained outstanding performance on different tasks, such as visual object recognition, image classification, hand-written character recognition and more. Deep CNNs introduced by LeCun et al. \cite{LeCun}, is a supervised learning model formed by multi-layer neural networks.
CNNs are fully data-driven and can retrieve hierarchical features automatically by building high-level features from low-level ones, thus obviating the need to manually customize hand-crafted features. Previous works have shown the benefit of using a fully convolutional architecture for liver lesion detection and segmentation applications \cite{Ben-Cohen,Christ}. Fully convolutional networks (FCN) can take input of arbitrary size and produce correspondingly-sized output with efficient inference and learning. Unlike patch based methods, the loss function using this architecture is computed over the entire image. The network processes entire images instead of patches, which removes the need to select representative patches, eliminates redundant calculations where patches overlap, and therefore scales up more efficiently with image resolution. Moreover, there is a fusion of different scales by adding links that combine the final prediction layer with lower layers with finer strides.

\subsection{Conditional Adversarial Networks}
\begin{figure}
\centering
\includegraphics[height=5.1 cm]{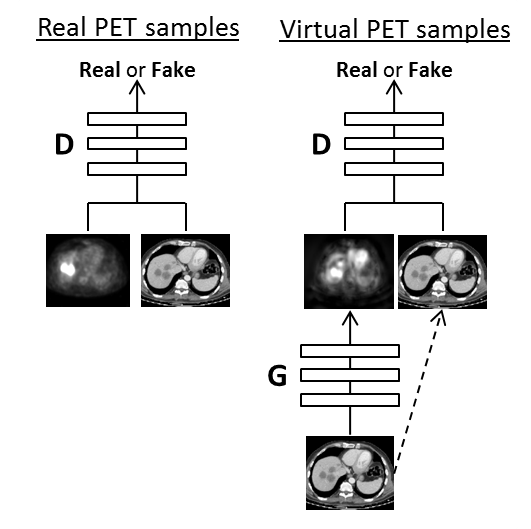}
\caption{Training a conditional GAN to predict PET images from
CT images. The discriminator, D, learns to classify between real and
synthesized pairs. The generator learns to fool the discriminator.}
\label{fig:gan}
\end{figure}

More recent works show the use of Generative Adversarial Networks (GANs) for image to image translation \cite{Isola}. GANs are generative models that learn a mapping from random noise vector z to output image y \cite{Goodfellow}. In contrast, conditional GANs learn a mapping from observed image x and random noise vector z, to y. The generator G is trained to produce outputs that cannot be distinguished from “real” images by an adversarially trained discriminator, D, which is trained to do the best possible to detect the generator’s “fakes”. Fig. \ref{fig:gan} shows a diagram of this procedure.
\\

In this study we explore FCN and conditional GAN for estimating PET-like images from CT volumes. The advantages of each method are used to create a realistic looking virtual PET images with specific attention to hepatic malignant tumors. To the best of our knowledge, this is the first work that explores CT to PET translation using deep learning.

\section{Methods}
\begin{figure}
\centering
\includegraphics[height=7.1 cm]{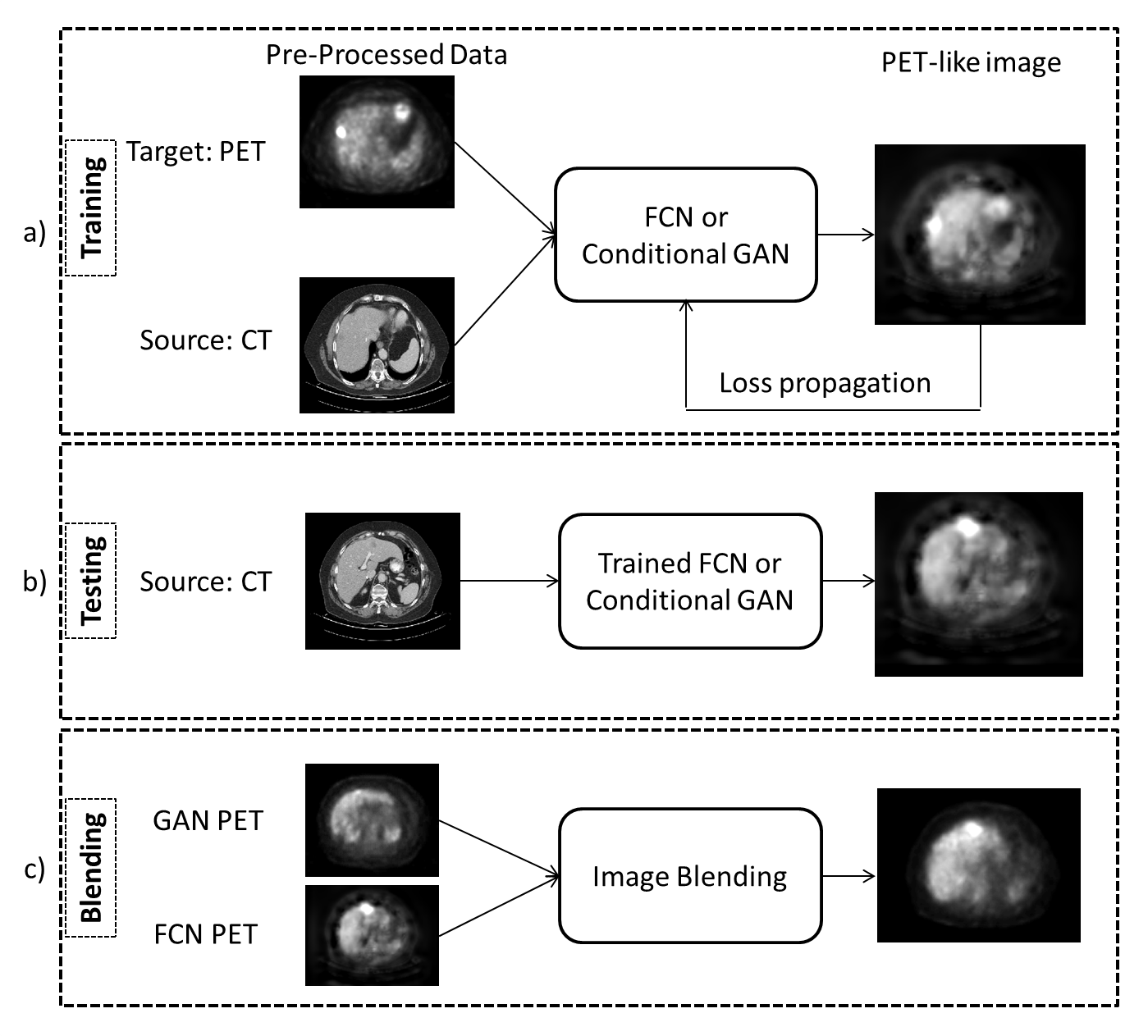}
\caption{The proposed virtual PET system.}
\label{fig:diagram}
\end{figure}

Our framework includes three main modules: training module which includes the data preparation; testing module which accepts CT images as input and predicts the virtual PET image as output; blending module which blends the FCN and the conditional GANs output. The FCN and conditional GANs play the same role for training and testing. We explore and use both of them for the task of predicting PET-like images from CT images. Fig. \ref{fig:diagram} shows a diagram of our general framework. Each module will be described in depth in the following subsections.

\subsection{Training Data Preparation}
The training input for the FCN or conditional GANs are two image types: source image (CT image) and target image (PET image) which should have identical size in our framework.
Hence, the first step in preparing the data for training was aligning the PET scans with the CT scans using the given offset, pixel-spacing and slice-thickness of both scans. Secondly, we wanted to limit our PET values to a constrained range of interest. The standardized uptake value (SUV) is commonly used as a relative measure of FDG uptake \cite{Higashi} as in equation \ref{eq:suv}:
\begin{equation} 
\label{eq:suv}
SUV=\frac{r}{a'/w}
\end{equation}

Where $r$ is the radioactivity concentration [kBq/ml] measured by the PET scanner within a region of interest (ROI), $a'$ is the decay-corrected amount of injected radiolabeled FDG [kBq], and $w$ is the weight of the patient [g], which is used as a surrogate for a distribution volume of tracer. The maximum SUV (termed SUVmax) was used for quantitative evaluation \cite{Kinehan}. 

Since the CT and PET scans include a large range of values, it makes it a difficult task for the network to learn the translation between these modalities and values range limitations were required. We used contrast adjustment, by clipping extreme values and scaling, to adjust the PET images into the SUV range of 0 to 20, this range includes most of the interesting SUV values to detect tumor malignancies. Similarly, CT images were adjusted into -160 to 240 HU as this is, usually, a standard windowing used by the radiologists.

\subsection{Fully Convolutional Network Architecture}
\begin{figure}
\centering
\includegraphics[height=7.1 cm]{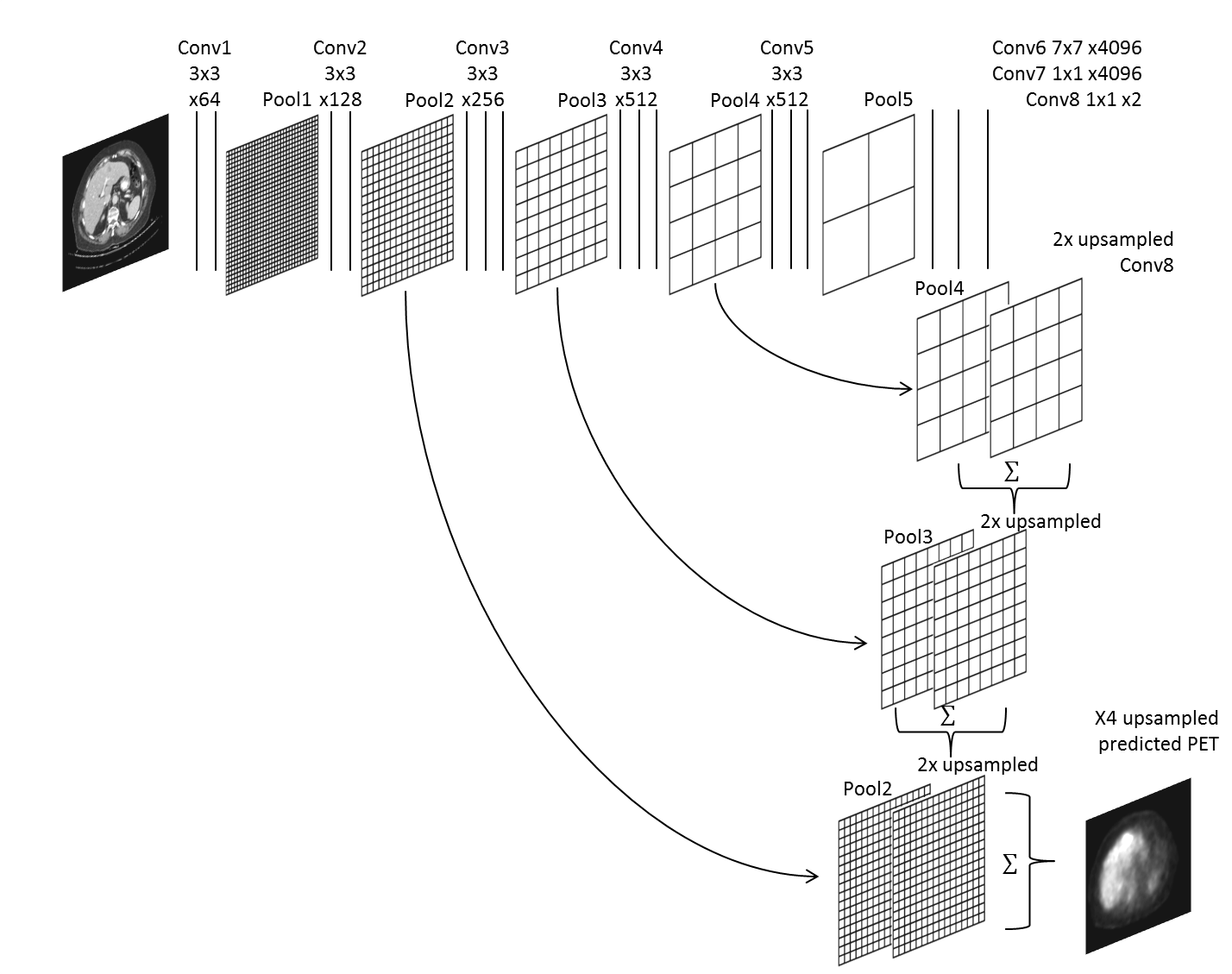}
\caption{FCN-4s architecture. Each convolution layer is illustrated by a straight line with the receptive field size and number of channels denoted above. The ReLU activation function and drop-out are not shown for brevity.}
\label{fig:FCN}
\end{figure}

In the following we describe the FCN used for both training and testing as in Fig. \ref{fig:diagram}a and \ref{fig:diagram}b. Our network architecture uses the VGG 16- layer net \cite{Simonyan}. We decapitate the net by discarding the final classifier layer, and convert all fully connected layers to convolutions. We append a 1x1 convolution with channel dimension to predict the PET images. Upsampling is performed in-network for end-to-end learning by backpropagation from the pixelwise $L_2$ loss. The FCN-4s net was used as our network, which learned to combine coarse, high layer information with fine, low layer information as described in \cite{Shelhamer} with an additional skip connection by linking the Pool2 layer in a similar way to the linking of the Pool3 and Pool4 layers in Fig. \ref{fig:FCN}.

\subsection{Conditional GAN Architecture}
Conditional GAN were used in a similar way described for the FCN in training and testing as in Fig. \ref{fig:diagram}a and \ref{fig:diagram}b. We adapt the conditional GAN architecture from the one presented in \cite{Isola}. The generator in this architecture is ``U-Net” based \cite{Ronneberger}. For the discriminator a ``PatchGan" classifier \cite{Isola} was used which only penalizes structure at the scale of image patches. Using a ``PatchGan" the discriminator tries to classify if each $70 \times 70$ patch in the image is real or fake.
Let $C_k$ denote a Convolution-BatchNorm-ReLU layer with $k$ filters and $CD_k$ denotes a Convolution-BatchNorm-Dropout-ReLU
layer with a dropout rate of 50\%. All convolutions are $4\times 4$ spatial filters. Convolutions
in the ``U-Net" encoder, and in the discriminator (except of its last convolution layer), downsample by a factor of 2, whereas in the ``U-Net" decoder they upsample by a factor of 2.
\\
For the conditional GAN we used the following architecture:
\begin{itemize}
\item The discriminator: $C_{64}-C_{128}-C_{256}-C_{512}-C_{1}$.
\item The ``U-Net" encoder: $C_{64}-C_{128}-C_{256}-C_{512}-C_{512}-C_{512}-C_{512}-C_{512}$. 
\item The ``U-Net" decoder: $CD_{512}-CD_{512}-CD_{512}-C_{512}-C_{512}-C_{256}-C_{128}-C_{64}$
\end{itemize}
The ``U-Net" includes skip connections between each layer $i$ in the encoder and layer $n-i$ in the decoder, where $n$ is the total number of layers. The skip connections concatenate activations from layer $i$ to layer $n-i$.

The ``U-Net" generator is tasked to not only fool the discriminator but also to be similar to the real PET image in an $L_2$ sense, similar to the regression conducted in the FCN. For additional implementation details please refer to \cite{Isola}.

\subsection{Loss Weights}
Our study concentrates on the malignant tumors in PET scans. Malignant tumors are usually observed with high SUV values ($>$2.5) in PET scans. Hence, we used the SUV value in each pixel as a weight for the pixel-wise loss function. By this we allow the network to pay more attention to high SUV value even though most pixels include low values.

\subsection{Image Blending}
Since the conditional GAN learns to create realistic looking images its output was much more similar to real PET than that of the FCN that provided blurred images. However, the FCN based system had much better response to malignant tumors compared to the conditional GAN. Hence we used the advantages of each method to create a blended image that includes the realistic looking images of the conditional GAN together with the more accurate response for malignant tumors using the FCN as in Fig. \ref{fig:diagram}c. 
First, we created a mask from the FCN output which includes regions with high predicted SUV values ($>$2.5). This mask marks the regions in which the FCN image will be used, where the rest of the image will include the conditional GAN image. A pyramid based blending was used \cite{Adelson}. Laplacian pyramids were built for each image and a Gaussian pyramid was built for the mask. The Laplacian pyramids were combined using the mask's Gaussian pyramid as weights and collapsed to get the final blended image.

\section{Results}

\subsection{Dataset}
The data used in this work includes CT scans with their corresponding PET scans from the Sheba Medical Center. The dataset contains 25 CT and PET pairs which we constrained to the region of the liver for our study.  Not all PET/CT scans in our dataset included liver tumors. The training set included 17 PET/CT pairs and the testing was performed on 8 pairs.

\subsection{Preliminary Results}
The generated virtual PET image, per input CT scan, was  visually evaluated by a radiologist.  The  virtual PET result was then compared to the real PET images by comparing tumor detection  in the liver region. We define a detected tumor as a tumor that has SUVmax value greater than 2.5. Two evaluation measurements were computed, the true positive rate (TPR) and false positive rate (FPR) for each case as follows:
\begin{itemize}
\item $TPR$- Number of correctly detected tumors divided by the total number of tumors.
\item $FPR$- Number of false positives per scan.
\end{itemize}

\begin{figure}
\centering
\includegraphics[height=12.0 cm]{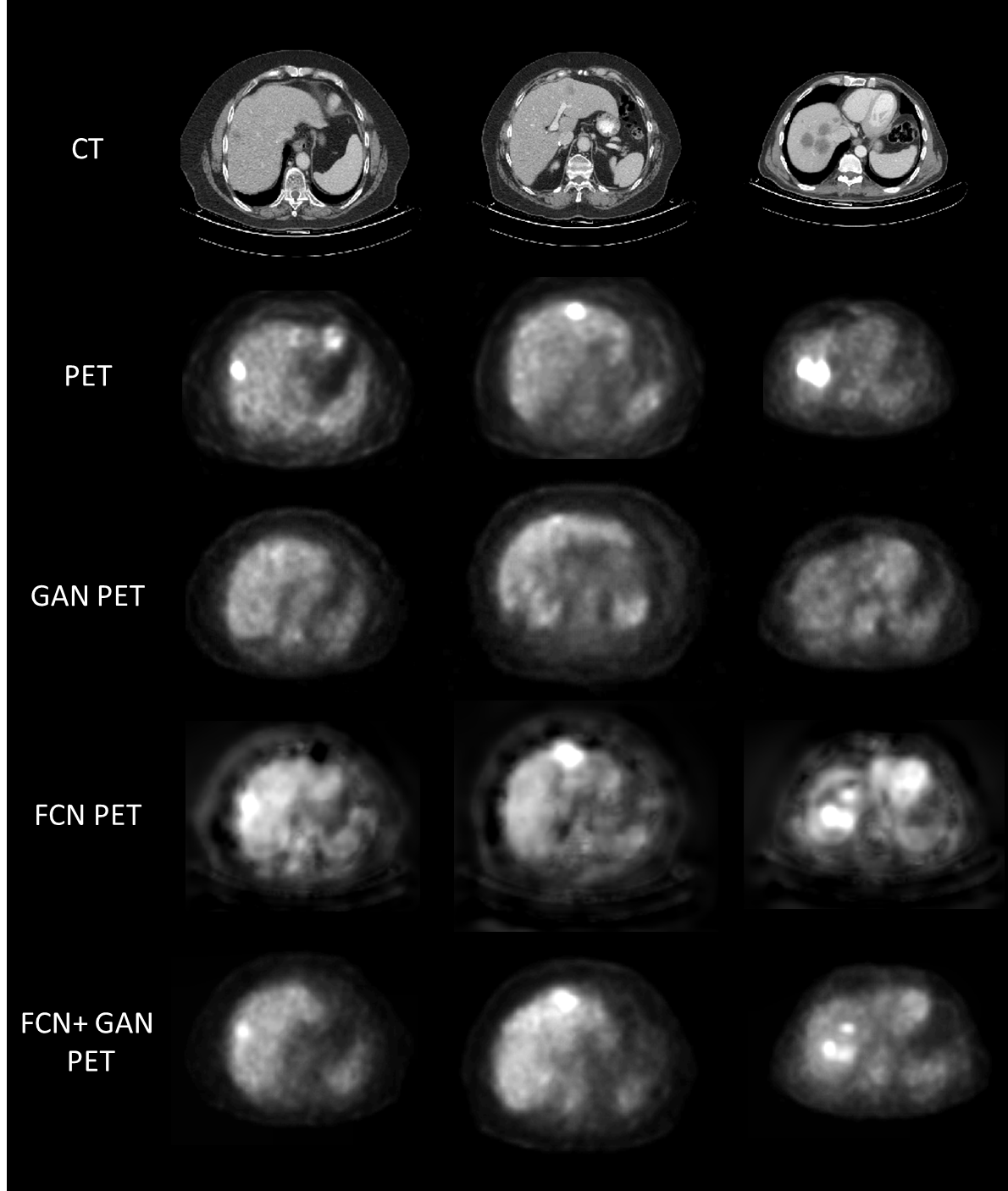}

\caption{Sample results of the predicted PET using FCN and conditional GAN compared to the real PET.}
\label{fig:results}
\end{figure}

The testing set included 8 CT scans with a total of 26 liver tumors. The corresponding PET scans were used as comparison with the predicted virtual PET. Our FCN and GANs based system successfully detected 24 out of 26 tumors (TPR of 92.3\%) with only 2 false positives for all 8 scans (average FPR of 0.25).

\begin{figure}
\centering
\includegraphics[height=3.0 cm]{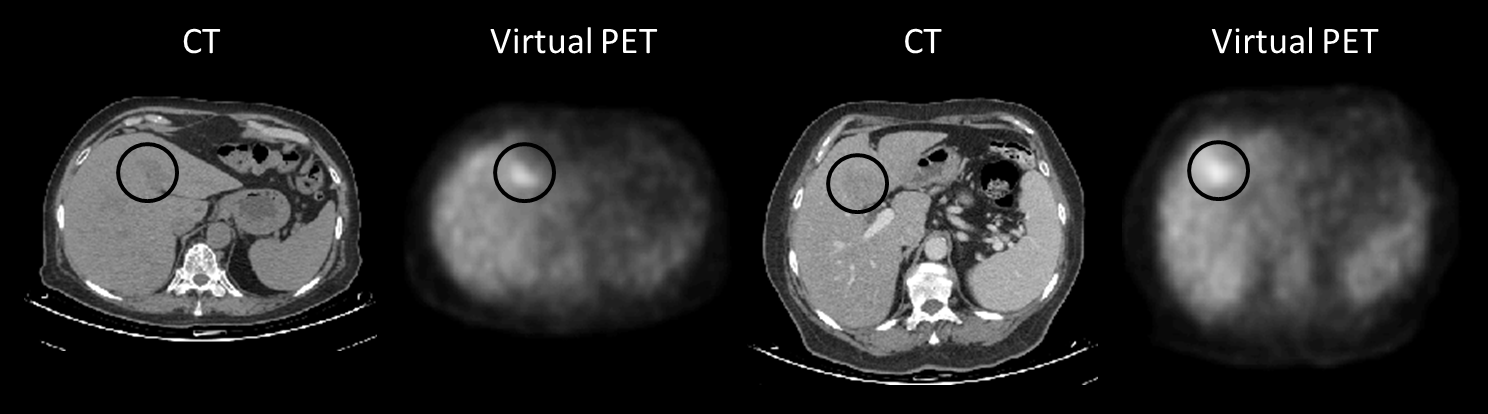}
\caption{False positive examples are marked with a black circle.}
\label{fig:FPs}
\end{figure}

Fig. \ref{fig:results} shows sample results obtained using the FCN, and FCN blended with the conditional GAN, compared to the real PET scan.
False positive examples are shown in Fig. \ref{fig:FPs}. In these cases, the FCN mistranslated hypodense regions in the liver to high SUV values.

\section{Conclusions}
A novel system for PET estimation using only CT scans has been presented. Using the FCN with weighted regression loss together with the realistic looking images of the conditional GAN our virtual PET results look promising detecting most of the malignant tumors which were noted in the real PET with a very small amount of false positives. In comparison to the FCN the conditional GAN did not detect the tumors but obtained images which were very similar to real PET. A combination of both methods improved the FCN output blurred appearance.
Future work entails obtaining a larger dataset with vast experiments using the entire CT and not just the liver region. The presented system can be used for many applications in which PET examination is needed such as evaluation of drug therapies and detection of malignant tumors.

\subsubsection*{Acknowledgment}
This research was supported by the Israel Science Foundation (grant No. 1918/16).

Part of this work was funded by the INTEL Collaborative Research Institute for Computational Intelligence (ICRI-CI).

Avi Ben-Cohen's scholarship was funded by the Buchmann Scholarships Fund.

\end{document}